\documentclass[10pt,twocolumn,letterpaper]{article}

\usepackage[pagenumbers]{cvpr} % To force page numbers, e.g. for an arXiv version

\usepackage[dvipsnames]{xcolor}

\definecolor{cvprblue}{rgb}{0.21,0.49,0.74}
\usepackage[pagebackref,breaklinks,colorlinks,citecolor=cvprblue]{hyperref}

\def\paperID{*****} % *** Enter the Paper ID here
\def\confName{CVPR}
\def\confYear{2024}
\usepackage{amsmath,bm}
\usepackage{gensymb}
\usepackage{booktabs}
\usepackage{amssymb}
\usepackage{pifont} 
\usepackage{algorithm}
\usepackage{algorithmicx}  
\usepackage{algpseudocode}  

\usepackage{fontawesome}
\usepackage{bbding}
\newcommand{\cmark}{\textcolor{green}{\ding{51}}} % 绿色对号
\newcommand{\xmark}{\textcolor{red}{\ding{55}}}   % 红色错号

\title{Cycle3D: High-quality and Consistent Image-to-3D Generation via Generation-Reconstruction Cycle}

\author{%
    Zhenyu Tang$^{1}$\thanks{Equal contribution.} \quad 
    Junwu Zhang$^{1}$\footnotemark[1] \quad
    Xinhua Cheng$^{1}$ \quad
    Wangbo Yu$^{1}$ \quad
    Chaoran Feng $^{1}$ \quad\\
    Yatian Pang$^{{1},{3}}$ \quad 
    Bin Lin $^{1}$ \quad
    Li Yuan$^{{1},{2}}$\thanks{Corresponding author.} \\
    $^{1}$Peking University 
    $^{2}$ Pengcheng Laboratory
    $^{3}$ National University of Singapore
}

\begin{document}
\maketitle
\begin{abstract}
% Recent 3D large reconstruction models typically employ a two-stage process: first generate multi-view images by a multi-view diffusion model, and then utilize a feed-forward model to reconstruct images to 3D content.
Recent 3D large reconstruction models typically employ a two-stage process, including first generate multi-view images by a multi-view diffusion model, and then utilize a feed-forward model to reconstruct images to 3D content. 
%and then utilize a feed-forward reconstruction model to transform these images to 3D content.  
However, multi-view diffusion models often produce low-quality and inconsistent images, adversely affecting the quality of the final 3D reconstruction. 
% Consequently, directly concatenating them for inference adversely affects the quality of 3D generation. 
To address this issue, we propose a unified 3D generation framework called \textbf{Cycle3D}, 
% which cascades the powerful pre-trained 2D diffusion model with the 3D reconstruction module into a unified diffusion process. 
which cyclically utilizes a 2D diffusion-based generation module and a feed-forward 3D reconstruction module during the multi-step diffusion process.
% \cxh{by turns for creating consistent 3D content}.
Concretely, 
% 2D diffusion model achieves high-quality generation 
2D diffusion model is applied for generating high-quality texture, 
% while the reconstruction model ensures 3D consistency 
and the reconstruction model guarantees multi-view consistency. 
% These two cascaded modules perform unified denoising at each step, achieving high-quality and consistent generation. 
% During the denoising process, the 2D diffusion model can also control the generation of unseen views and inject reference-view information, thereby enhancing the diversity and texture consistency of 3D generation.
Moreover, 2D diffusion model can further control the generated content and inject reference-view information for unseen views, thereby enhancing the diversity and texture consistency of 3D generation during the denoising process.
Extensive experiments demonstrate the superior ability of our method to create 3D content with high-quality and consistency compared with state-of-the-art baselines. 
Our project page is available at \url{https://pku-yuangroup.github.io/Cycle3D/}.
\end{abstract}    
\section{Introduction}
\label{sec:intro}
The presence of high-quality and diverse 3D assets is essential across various fields, such as robotics, gaming, and architecture. Traditionally, the creation of these assets has been a labor-intensive manual process, necessitating proficiency with complex computer graphics software. Consequently, the automatic generation of diverse and high-quality 3D content from single-view images has emerged as a crucial objective in 3D computer vision. 

% Previous research~\cite{qian2023magic123, tang2023make, sun2023dreamcraft3d, wu2024consistent3d,wang2023prolificdreamer,yi2024diffusion} on 3D generation has explored distilling 2D diffusion priors into 3D representations through score distillation sampling~\cite{poole2022dreamfusion}. However, these methods often require several hours to generate a single 3D object, making them impractical for real-world applications.
\begin{figure}[!t]
    \centering
    \includegraphics[width=0.45\textwidth]{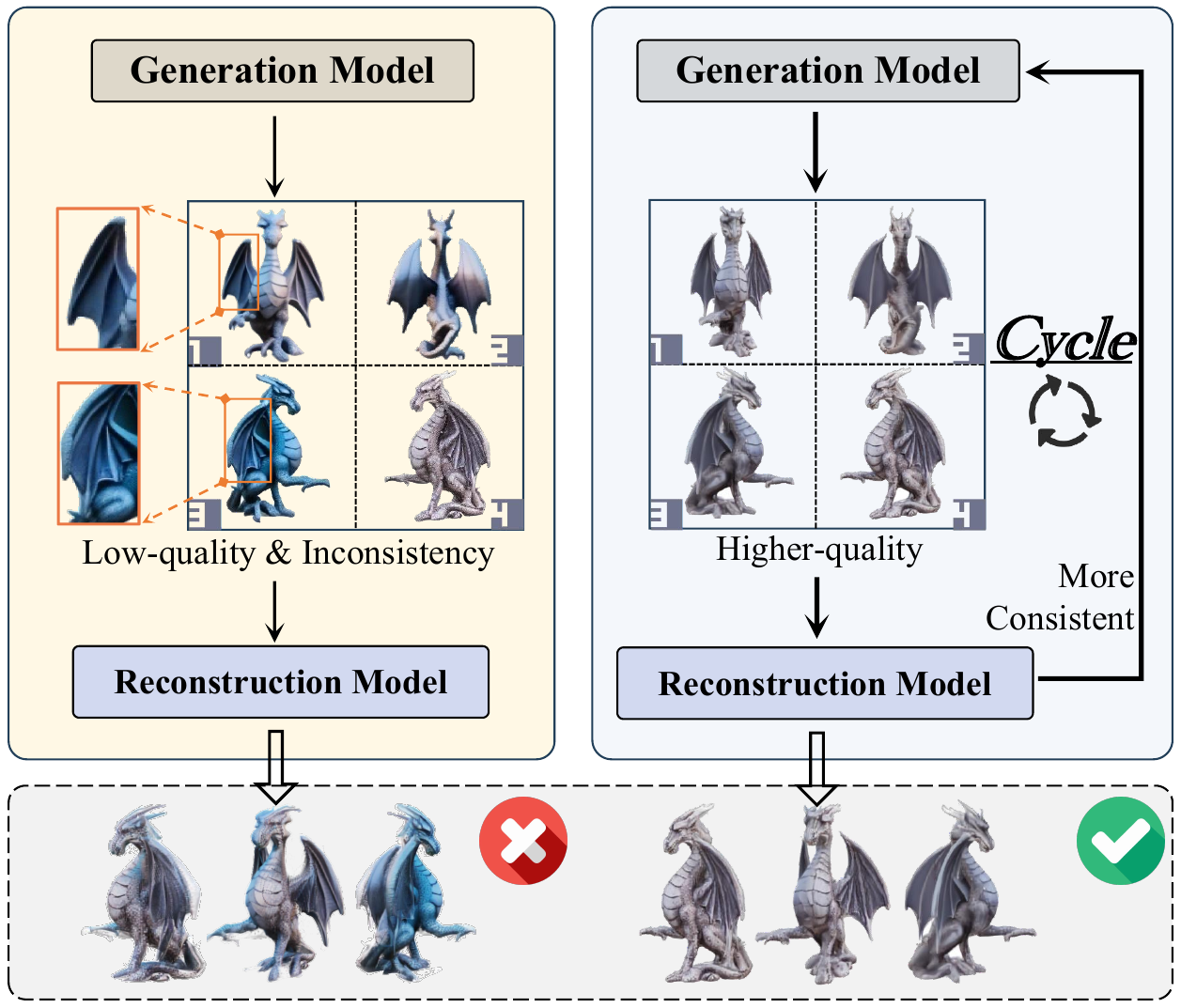}
    \caption{\textbf{Motivation of our pipeline.} Current large-scale reconstruction models often produce geometric artifacts and blurry textures due to the limited quality and consistency of the multi-view images generated by multi-view diffusion models. 
    Our Cycle3D cyclically uses a 2D diffusion-based generation model and reconstruction model during the multi-step diffusion process. During denoising, 2D generation model improves image quality, while the reconstruction model enhances 3D consistency.}
    \label{fig:motivation}
    
\end{figure}

With the emergence of large-scale 3D datasets~\cite{deitke2023objaverse,deitke2024objaversexl,yu2023mvimgnet,wu2023omniobject3d}, recent research~\cite{xu2024instantmesh,wei2024meshlrm,li2023instant3d,wang2024crm,xu2024grm,tang2024lgm} has focused on large 3D reconstruction models. These models typically combine multi-view diffusion models and sparse-view reconstruction models to directly predict 3D representations (Triplane-NeRF~\cite{shue20233d,chan2022efficient}, and 3D Gaussian Splatting~\cite{3dgs}), enabling efficient 3D generation in a feed-forward manner. 
\begin{figure*}
    \centering
    \includegraphics[width=0.955\textwidth]{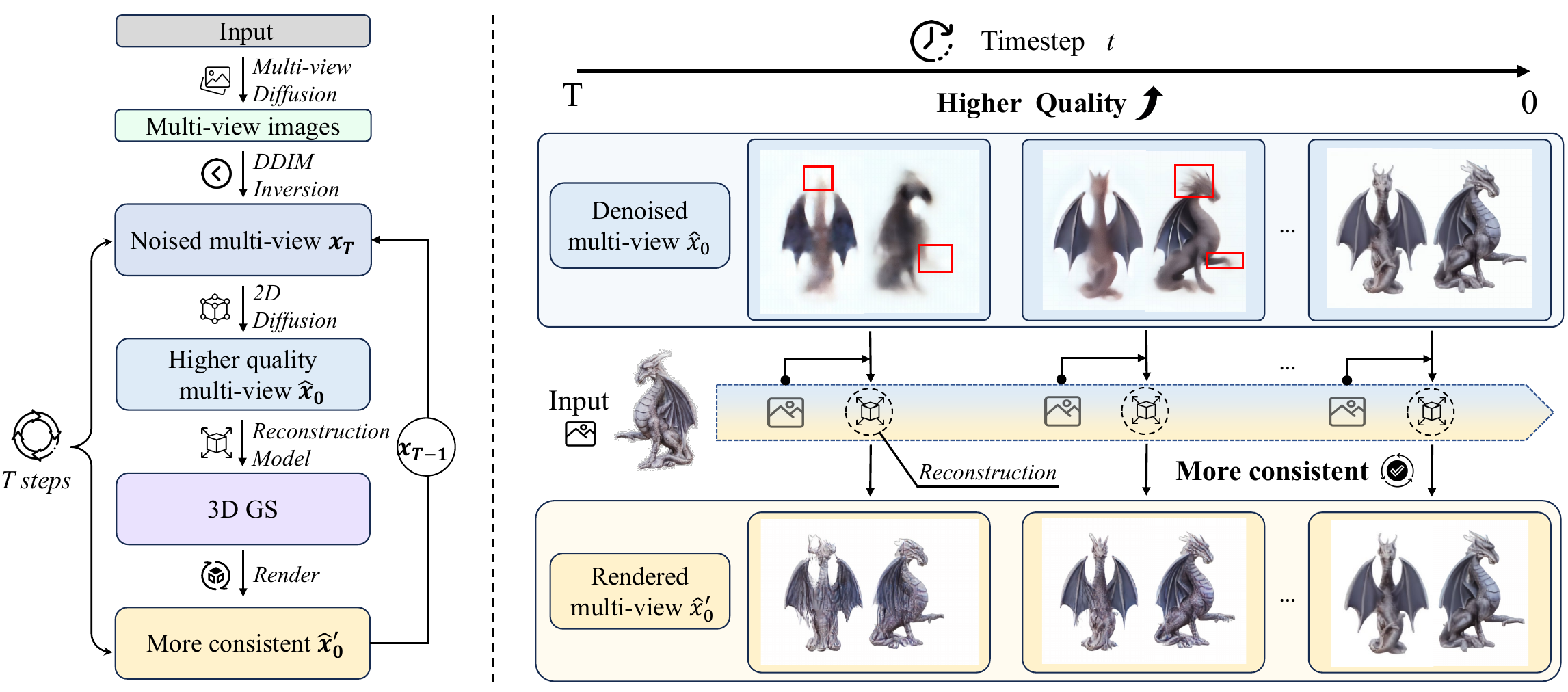}
    \caption{\textbf{Overview of our Cycle3D. }The left side illustrates the Cycle3D workflow, while the right side visualizes the denoising process. During the multi-step denoising process, the input view remains clean, the pre-trained 2D generation model gradually produces multi-view images with higher quality, while the reconstruction model continuously corrects their 3D inconsistencies. The red boxes highlight inconsistencies between the multi-view images, which are then corrected by reconstruction model.}
    \label{fig:motivation2}
\end{figure*}

However, we have observed that existing methods often encounter following two issues as shown in Figure~\ref{fig:motivation}: (1) \textbf{Low quality}: Multi-view diffusion models and reconstruction models are trained on limited synthetic 3D datasets, resulting in low-quality generation and poor generalization to real-world scenarios. (2) \textbf{Multi-view inconsistency}: Multi-view diffusion models struggle to generate pixel-level consistent multi-view images, while reconstruction models are typically trained on consistent ground truth multi-view images. Consequently, inconsistent multi-view images usually significantly affect reconstruction results, leading to geometric artifacts and blurry textures.

To address these challenges, In this paper, we propose \textbf{Cycle3D}. Our method is designed based on the following two key insights: (1) The pre-trained 2D diffusion model trained on billions of web images can generate high-quality images, which is beneficial to 3D reconstruction; (2) 
% The reconstruction model can ensure 3D consistency across multi-views in 2D diffusion. 
The reconstruction model can ensure consistency across multi-views and inject consistency in 2D diffusion generation. Specifically, as shown in Figure~\ref{fig:motivation2}, we propose a unified image-to-3D generation framework that cyclically utilizes a pre-trained 2D diffusion model and a feed-forward 3D reconstruction model during multi-step diffusion process.
First, we inverse the multi-view images generated by multi-view diffusion into the initial noise, serving as shape and texture priors. Then, in each denoising step, multi-view images are denoised and reconstructed to 3D-GS to be re-rendered, forming a loop to continue multi-step denoising. During the denoising process, the 2D diffusion model gradually provides higher quality multi-view images, while the reconstruction module progressively corrects 3D inconsistencies across multi-views. The reconstruction model can further enhance the reconstruction quality through interaction with features in the 2D Diffusion model. Additionally, 2D diffusion can control the generation of unseen views and inject reference view information during the denoising process due to the advanced development, which further enhances the diversity and consistency of 3D generation.

We conducted extensive qualitative and quantitative experiments to validate the efficacy of our proposed Cycle3D. The experimental results demonstrate that Cycle3D outperforms other feed-forward methods and even surpasses some optimization-based methods on image-to-3D tasks.  In summary, Our main contributions can be summarized as follows:
\begin{itemize}
\item 
We propose a unified image-to-3D generation framework, \textbf{Cycle3D}, which cyclically uses 2D diffusion model and a 3D reconstruction model during multi-step diffusion process. In this framework, 2D diffusion model improves the quality of multi-view images, and the reconstruction model enhances 3D consistency. The feature interaction between 2D diffusion and reconstruction model further improves the reconstruction quality.

\item
Leveraging the 2D diffusion model, Cycle3D can control the generation of unseen views and inject reference-view information, thereby enhancing the diversity and texture consistency of 3D generation.

\item 
Our experiments demonstrate that our framework surpasses existing methods, achieving satisfactory image-to-3D generation with high-quality and 3D consistency.

\end{itemize}

\section{Related Works}
\textbf{3D Generation from one image}
3D generation from a single image is a crucial task in computer vision, which is mainly divided into two approaches: (1) Optimization-Based Methods:
These methods optimize 3D representation using 2D or multi-view diffusion models for Score Distillation Sampling  sampling (SDS)~\cite{poole2022dreamfusion, tang2023make, qian2023magic123, tang2023dreamgaussian, zhang2023repaint123fasthighqualityimage, yu2024hifi123highfidelityimage3d, evagaussians, cheng2023progressive3d,yang2024fourier123, huang2023dreamtime}. They iteratively optimize the 3D representation but are computationally intensive, leading to long optimization time.
(2) Feed-Forward Generation Methods:
These methods generate 3D models in a single forward pass, offering faster generation speed~\cite{tang2024lgm, hong2023lrm, xu2024grm, wang2024crm, li2023instant3d, xu2024instantmesh, jiang2023leapliberatesparseview3d}. These methods provide a quicker alternative to optimization-based approaches, balancing speed and quality. Our work also involves generating high-quality and consistent 3D models in the feed-forward manner. %which takes only 25 seconds.

\textbf{Large Reconstruction Model for 3D Generation.}
These methods~\cite{tang2024lgm, hong2023lrm, xu2024grm, wang2024crm, li2023instant3d, xu2024instantmesh, jiang2023leapliberatesparseview3d} typically involve multi-view diffusion models~\cite{shi2023mvdream, wang2023imagedreamimagepromptmultiviewdiffusion} to generate multi-view images, followed by the feed-forward reconstruction model to obtain the 3D representation. LGM~\cite{tang2024lgm} uses U-Net as the 3D reconstruction model, while LRM~\cite{hong2023lrm} and GRM~\cite{xu2024grm} employs transformers. The generative capability mainly stems from multi-view diffusion model, with large reconstruction model primarily focusing on faithful 3D reconstruction. However, multi-view diffusion model cannot guarantee 3D consistency, leading to reconstruction artifacts. Our method uses a multi-step approach for jointly generation and reconstruction, employing the reconstruction model to continuously correct inconsistency and 2D diffusion to progressively enhance quality, resulting in high-quality, consistent 3D models.
\section{Preliminary: Gaussian Splatting}
\begin{figure*}
    \centering
    \includegraphics[width=0.86\textwidth]{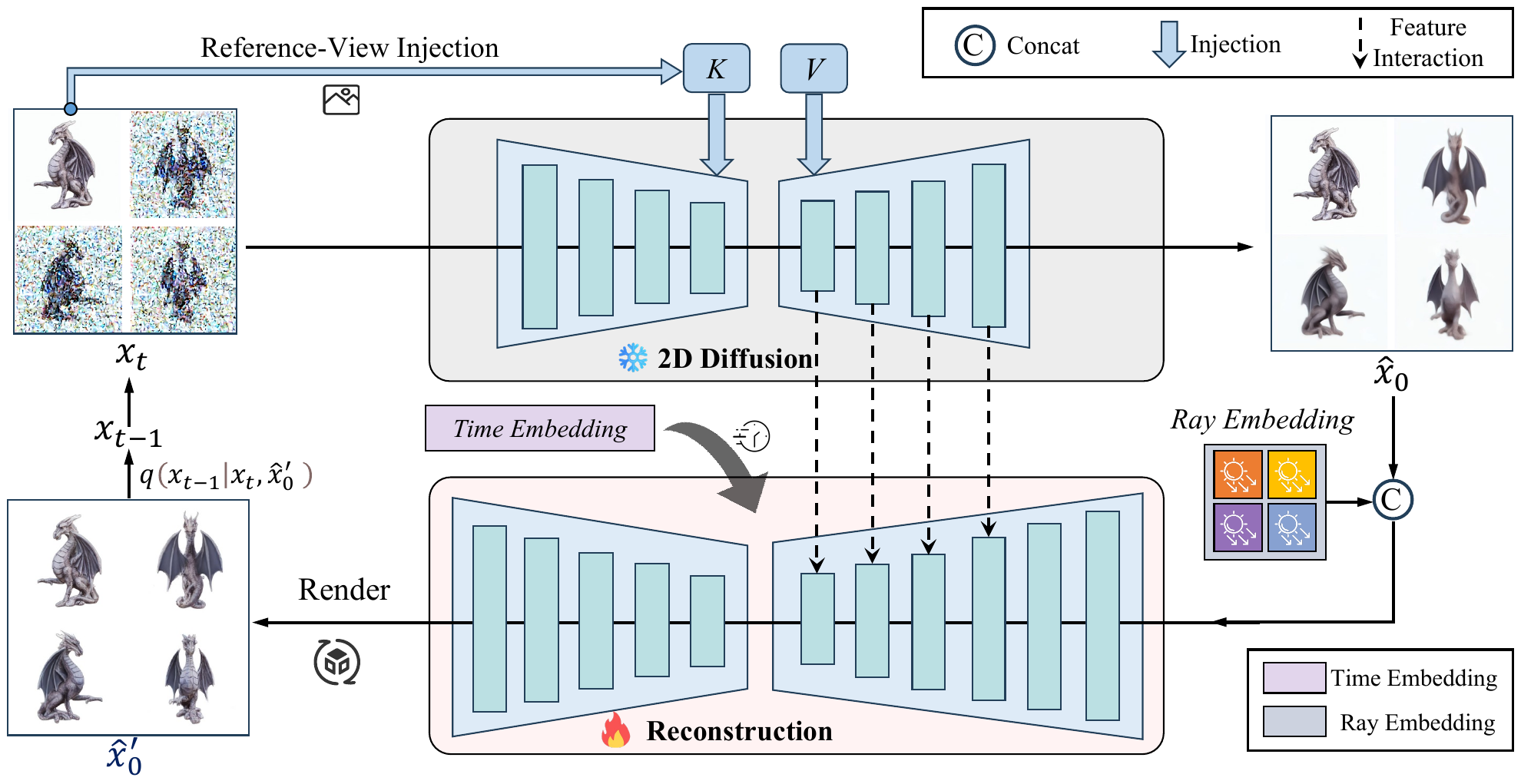}
    \caption{\textbf{Process of our Cycle3D. }We propose a unified image-to-3D Diffusion framework that cyclically utilizes pre-trained 2D Diffusion model and 3D reconstruction model. During denoising, 2D Diffusion model can inject reference-view features, and the reconstruction model incorporates time embeddings to adapt to  \( \mathbf{\hat{x}}_0 \)  at different timesteps. Additionally, the interaction between features of reconstruction model's encoder and 2D Diffusion model's decoder enhances robustness of reconstruction. During inference, we use the multi-view images \(\mathbf{ \hat{x}}'_0 \)  rendered by reconstruction model and the previous step \(\mathbf{x}_t\) , resampling to obtain \(\mathbf{x}_{t-1}\), while keeping the reference view clean.}
    \label{fig:main}
\end{figure*}

Gaussian Splatting~\cite{3dgs} introduces an innovative approach for synthesizing new views and fitting 3D scenes, achieving real-time performance. Gaussian Splatting employs a set of anisotropic 3D Gaussians, to represent the scene. Specifically, each Gaussian is composed of its 3D position $p \in \mathbb{R}^3$,  3D scale $s \in \mathbb{R}^3$ or 2D scale $s \in \mathbb{R}^2$, color $c \in \mathbb{R}^3$, opacity $\alpha \in \mathbb{R}$, and a rotation quaternion $q \in \mathbb{R}^4$.  These 3D Gaussians are projected onto the image plane as 2D Gaussians and rendered in real
time via the tiled rasterizer.

\section{Method}
\label{sec:method}

Given an RGB image, Cycle3D aims to generate high-quality and consistent 3D objects using diffusion model and reconstruction model. Specifically, as illustrated in Figure~\ref{fig:main}, our framework utilizes a pre-trained frozen 2D diffusion model(refer to Sec~\ref{sec41}) to denoise multi-view images and a reconstruction model(refer to Sec~\ref{sec42}) to correct inconsistencies and reconstruct 3D content. 
% This allows us to jointly train the cascade diffusion model end-to-end. 
Then, we cascade the 2D generation model and 3D reconstruction model in a unified diffusion process and perform generation-reconstruction cycle(refer to Sec~\ref{sec43}) denoising to achieve high-quality and consistent 3D results.

\subsection{Pre-trained 2D Diffusion}
\label{sec41}
Recent image-based multi-view diffusion models~\cite{kim2024multi, zuo2024videomv} are usually trained on limited synthetic 3D data, which hinders their ability to capture fine texture details and generalize to real-world scenarios. Therefore, we employ a 2D diffusion model~\cite{Rombach_2022_CVPR} (Stable Diffusion 1.5) that has been pre-trained on a large number of web images to generate high-quality multi-view images. Specifically, during inference, we first use the multi-view diffusion model~\cite{kim2024multi} to obtain multi-view as the basic shape prior, then inverse multi-view images to noise by performing DDIM~\cite{song2020denoising} . The 2D diffusion model, through class-free guidance denoising, improves the quality of multi-view generation and enhances texture details. During the denoising process, to ensure consistency between the reference view and the condition image, we set the timestep of the input view to always be 0 and keep the condition image clean.

The 2D diffusion model effectively aligns with text prompts, so we can use more diverse text prompts to control the generation of regions not visible in the input view during the denoising process of multi-view images. Unlike directly using results generated by the multi-view diffusion model, our approach allows us to achieve more diverse 3D generation by using the customized text prompt.
Furthermore, benefiting from the advanced development of 2D diffusion technology,  we incorporate reference-view attention features into the diffusion denoising process inspired by~\cite{cao2023masactrl}. By concatenating attention keys and values between  non-input views and the reference view, we can obtain more consistent textures in multi-view images, improving the quality of image-to-3D generation. 

In our framework, the 2D diffusion model does not independently complete the entire denoising process. Within the denoising loop, we directly estimate $\hat{\mathbf{x}}_{0}$
from the noise predicted by the 2D Diffusion model at every timestep, which is then used for the following 3D reconstruction. We represent this process as follows:
\begin{align}
    \hat{\mathbf{x}}_{0} = \frac{1}{\sqrt{\bar{\alpha}_t}}( \mathbf{x}_t-\sqrt{1-\bar{\alpha}_t} \boldsymbol{\epsilon}_{\theta}(\mathbf{x}_t, y, t)).
    \label{eq:one-step-mvd} 
\end{align}
where $\boldsymbol{\epsilon}_{\theta}$  denotes 2D diffusion model,  $\alpha_t$ and $\bar{\alpha}_t$ schedules the amount of noise added at timestep $t$, with $y$ representing the text description.
Subsequently, we use the frozen VAE to transform $\hat{\mathbf{x}}_{0}$ from the latent space to the image space.
\subsection{ Reconstrcution Model}
\label{sec42}
In the Cycle3D framework, we utilize a feed-forward reconstruction model to predict attributes of 3D Gaussians from the multi-view 
$\hat{\mathbf{x}}_{0}$
obtained via 2D diffusion model, thereby recovering 3D models. Here, we employ an asymmetric U-Net Transformer $\mathcal{G}_{\phi}$ as proposed in~\cite{tang2024lgm}, which predicts pixel-aligned Gaussian parameters from the feature of each pixel in the final layer of the U-Net. Benefiting from the differentiable real-time rendering of Gaussian Splatting, reconstruction model can be integrated into our framework for end-to-end training, enabling efficient tuning.

In the denoising process of the 2D diffusion model, different timesteps produce varying levels of noise, which in turn affect the image qaulity of the directly estimated $\hat{\mathbf{x}}_{0}$. Therefore, to enhance the robustness of the model in reconstructing $\hat{\mathbf{x}}_{0}$
  at different timestep, we insert zero-initialized projection layers into each ResNetBlock within the U-Net. These layers map the time embeddings from the 2D Diffusion model to fit the reconstruction model. This creative adjustment helps the reconstruction model adapt to the $\hat{\mathbf{x}}_{0}$
  estimated at different timestep, thereby significantly enhancing the quality of 3D reconstruction.

% The limited availability of 3D datasets constrains the robustness of the reconstruction model when dealing with real-world images. The reconstruction process can be understood as a sequence from multi-view images to multi-view features and ultimately to multi-view Gaussians, Based on these observations, we propose enhancing the reconstruction process by leveraging features from a 2D Diffusion model pre-trained on a large number of web images. Specifically, we interact the decoder features from the 2D Diffusion model with the encoder features of the reconstruction model by introducing zero-initialized cross-attention layers within the reconstruction model. Through this innovative modification, the reconstruction model becomes more robust for reconstructing real-world images.

To further tune the reconstruction model to adapt to our enhancements, we supervise the training using T images $\boldsymbol{\hat I}$ and alpha masks $\boldsymbol{\hat M}$ rendered by $\mathcal{G}_{\phi}$ with the corresponding ground truth $\boldsymbol{I}$ and $\boldsymbol{M}$. The loss function is as followers:
\begin{align}
    &\mathcal{L}_{total} = \sum_{t=1}^{T} \left( \mathcal{L}_{\rm img}(\boldsymbol{\hat{I}_t}, \boldsymbol{I_t}) + ||\boldsymbol{\hat{M}_t} - \boldsymbol{M_t}||_2 \right), \label{eq:loss1}\\
    &\mathcal{L}_{\rm img}(\boldsymbol{\hat{I}_t}, \boldsymbol{I_t}) = ||\boldsymbol{\hat{I}_t} - \boldsymbol{I_t}||_2 + \lambda *\mathcal{L}_{\rm LPIPS}(\boldsymbol{\hat{I}_t}, \boldsymbol{I_t}),
     \label{eq:loss}
\end{align}
where $\mathcal{L}_{\rm LPIPS}$
is a perceptual image patch similarity loss~\cite{zhang2018unreasonable}, and the weight $\lambda$ is set to 0.50.
\subsection{Generation-Reconstruction Cycle}
\label{sec43}
\begin{table*}
\centering
\resizebox{0.8\textwidth}{!}{
\begin{tabular}{lccccc}
\toprule
          Methods/Metrics &    PSNR$\uparrow$ &   SSIM$\uparrow$ &  LPIPS$\downarrow$ &   CLIP-Similarity$\uparrow$ &  Contextual-Dis$\downarrow$ \\
\midrule
% \textit{\textbf{Optimization-Based Methods}}\\
    
   DreamGaussian~\cite{tang2023dreamgaussian}& 19.4900 & 0.8311 & 0.1145 & 0.7136 &      1.8139 \\
        Wonder3D~\cite{long2024wonder3d} & 18.0926 & 0.8164 & 0.1764 & 0.7596 &      1.7914 \\
        \midrule
% \textit{\textbf{Feed-Forward Based Methods}}\\
  One-2-3-45~\cite{liu2024one}& 14.0064 & 0.7405 & 0.3976 & 0.6363 &      2.1069 \\
TriplaneGaussian~\cite{zou2024triplane}& 18.4044 & 0.8284 & 0.1515 & 0.7399 &      1.7803 \\
         OpenLRM~\cite{hong2023lrm} &    18.6433 &    0.8301 &   0.1255 & 0.7567 &      1.7037 \\
         LGM~\cite{tang2024lgm} & 18.6909 & 0.8320 & 0.1417 & 0.7990 &  1.6504 \\
            Cycle3D (Ours) & \textbf{20.2452} & \textbf{0.8729} & \textbf{0.1117} & \textbf{0.8238} &      \textbf{1.6031} \\
\bottomrule
\end{tabular}
}
\caption{We show quantitative results of image-to-3D in terms of  PSNR$\uparrow$ / SSIM$\uparrow$, LPIPS$\downarrow$ / CLIP-Similarity$\uparrow$ / Contextual-Distance$\downarrow$ for our test dataset. The \textbf{bold} reflects the best result for optimization-based methods and feed-forward methods.}
\label{table:image-to-3D}
% \vspace{-0.6cm}
\end{table*}
The pre-trained 2D diffusion model exhibits powerful image generation capabilities but suffers from poor multi-view image consistency. In contrast, reconstruction model reconstructs multi-view images with accurate 3D consistency. Therefore, we developed a Generation-Reconstruction Cycle, cascading 2D diffusion and 3D reconstruction model into an iterative end-to-end pipeline, where 2D diffusion enhances quality and reconstruction model corrects inconsistencies. Instead of inputting the denoised latent \(\mathbf{x}_{t-1}\) into reconstruction model, we decode predicted \( \hat{\mathbf{x}}_0 \) using the VAE decoder to obtain clean multi-view images for reconstruction model. This aligns with the pretraining inputs of the reconstruction model, reducing the gap and easing the joint fine-tuning. Additionally, instead of using 2D diffusion model output \( \hat{\mathbf{x}}_0 \) to perform the DDIM backward step $ q(\mathbf{x}_{t-1} | \mathbf{x}_{t}, \hat{\mathbf{x}}_0)$ for updating \(\mathbf{x}_t\) to \(\mathbf{x}_{t-1}\), we adopt multi-view images \( \hat{\mathbf{x}}'_0 \) rasterized by 3D Gaussians output of the reconstruction model from the same observing views as \( \hat{\mathbf{x}}_0 \): 
\begin{align}
   \mathbf{x}_{t-1}=\frac{\sqrt{\alpha_{t-1}} \beta_{t}}{1-\bar{\alpha}_{t}} \hat{\mathbf{x}}'_0 +\frac{\sqrt{\alpha_{t}}\left(1-\bar{\alpha}_{t-1}\right)}{1-\bar{\alpha}_{t}} \mathbf{x}_{t},
\label{eq:sample}
\end{align}
where $\beta_{t} = 1 - \alpha_{t}$. Compared to \( \hat{\mathbf{x}}_0 \), \( \hat{\mathbf{x}}'_0 \) has accurate 3D consistency, making the sampling trajectory more 3D consistent.
Consequently, the final denoised result \(\mathbf{x}_0\) is more consistent and of higher quality than the result denoised only by the existing multi-view diffusion model, leading to the reconstruction of higher-quality 3D Gaussian structures.

Furthermore, training reconstruction model on a limited synthetic 3D dataset may affect its performance with real-world images. Based on the observation that reconstruction model reconstruction process can be understood as a sequence from multi-view images to multi-view features, and finally to multi-view Gaussians, we can enhance the reconstruction process using features from the 2D diffusion model pretrained on a large number of web images. Specifically, we introduce zero-initialized cross-attention layers in the reconstruction model to interact the decoder features of the 2D diffusion model with the encoder features of the reconstruction model, forming a U-Net structure. This innovative modification makes the reconstruction model more robust when reconstructing real-world images.
% \input{tables/performance1}
% \begin{figure*}
%     \centering
%     \includegraphics[width=0.88\textwidth]{fig/visualize.pdf}
%     \caption{Qualitative comparisons on image-to-3D generation. Zoom in for more details. }
%     \label{fig:main_fig}
% \end{figure*}

\section{Experiment}

\subsection{Implementation Details}
\label{details}
\textbf{Datasets} We use the G-objaverse dataset~\cite{qiu2024richdreamer} to train our model. Derived from the original Objaverse~\cite{deitke2023objaverse}, G-objaverse excludes 3D models with poor captions and includes a large number of high-quality renderings generated through a hybrid technique involving rasterization and path tracing. We utilize a further filtered subset containing approximately 80K 3D objects. Each model is rendered with 36 views, from which we randomly sample 4 views with elevation angles in the range [-5\degree, 5\degree] as input multi-views, using the first frame as the condition image. Additionally, we sample 8 views from the 36 views for extra supervision.

We collected real-world images and combined with those from the Realfusion15 dataset~\cite{melas2023realfusion} and the dataset collected by Make-It-3D~\cite{tang2023make}, using these images from diverse styles as the test dataset. Additionally, we further evaluate the 3D generation quality on 50 objects from the GSO dataset~\cite{downs2022google}, which were not included in the training set.

\textbf{Experimental Settings}
Our Cycle3D is trained on 8 NVIDIA A100(80G) with batch size 8 for about 1 day. We utilized the AdamW optimizer with a learning rate of 1e-4 and a weight decay of 0.05 for 30 epochs. Additionally, we followed ~\cite{tang2024lgm} to clip the gradient with a maximum norm of 1.0 and employed BF16 mixed precision with Deepspeed Zero2~\cite{rasley2020deepspeed}for efficient tuning. During inference, we use the DDIM scheduler, setting the sampling steps to 30, and take about 25 seconds to generate a 3D object.

\textbf{Evaluation Metrics}
We used PSNR, SSIM, and LPIPS~\cite{zhang2018unreasonable} to measure reconstruction quality, and CLIP score~\cite{radford2021learning} and contextual distance~\cite{mechrez2018contextual} to assess image similarity. The quality of 3D generation was evaluated by comparing 180 rendered views with the ground truth.

\textbf{Baselines} We select baselines for comparison, including optimization-based state-of-the-art image-to-3D methods: DreamGaussian~\cite{tang2023dreamgaussian}, Wonder3D~\cite{long2024wonder3d}, and some existing feed-forward methods:  One-2-3-45~\cite{liu2024one}, TriplaneGaussian~\cite{zou2024triplane}, LRM~\cite{hong2023lrm}, LGM~\cite{tang2024lgm}.

\begin{figure*}[!t]
    \centering
    \includegraphics[width=0.9\textwidth]{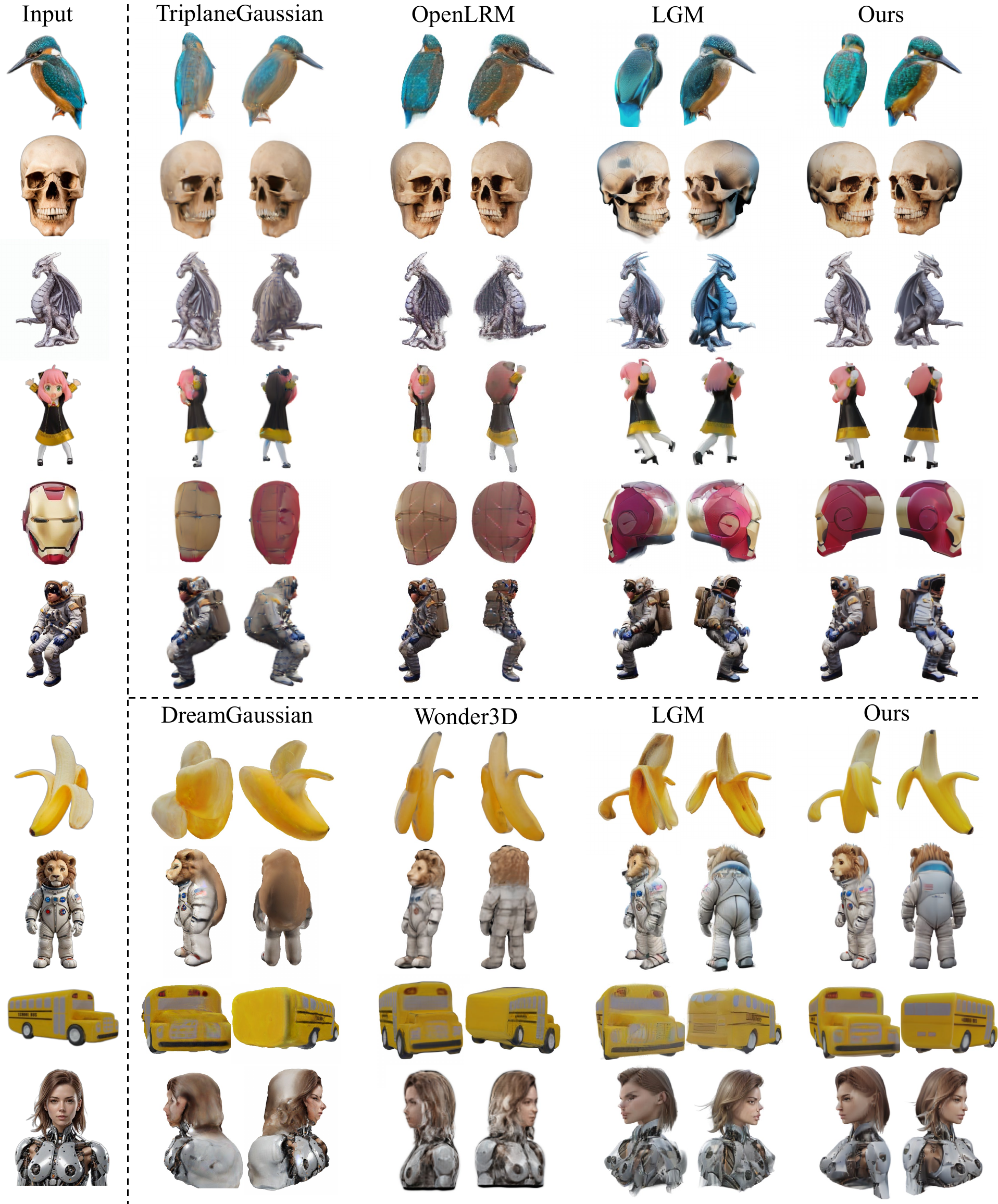}
    \caption{Qualitative comparisons on image-to-3D generation. Zoom in for more details. }
    \label{fig:main_fig}
\end{figure*}

\subsection{Comparison}

\textbf{Qualitative Comparisons.} We compared our approach with recent optimization-based and feed-forward based methods. For
more fair comparison, Cycle3D and LGM take the same generated multi-view inputs. As shown in Figure~\ref{fig:main_fig}, we used a wide range of wild images to evaluate the quality of image-to-3D generation, and our Cycle3D achieved the best visual results. TriplaneGaussian~\cite{zou2024triplane} and OpenLRM~\cite{hong2023lrm} fail to complete unseen regions with high quality. DreamGaussian~\cite{tang2023dreamgaussian} often produces unrealistic geometry, while Wonder3D~\cite{long2024wonder3d} tends to generate blurry textures.
LGM~\cite{tang2024lgm} often generates blurry textures and geometric artifacts like floating 3D Gaussian splats, due to low-quality and inconsistent multi-view images. In contrast, our method can generate high-quality and consistent 3D objects due to the cascaded diffusion process.

\textbf{Quantitative Comparisons.} As presented in Table~\ref{table:image-to-3D}, we quantitatively evaluate the quality of 3D generated objects for our test dataset. Notably, Cycle3D surpasses all baselines on all metrics, even outperforming existing optimization-based methods.
Furthermore, we validate our superiority on the GSO dataset~~\cite{downs2022google}, as shown in Table~\ref{table:gso}.
\begin{table}
\centering
\resizebox{0.485\textwidth}{!}{
\begin{tabular}{ccccccc}
        \toprule
        (I) & (II) & PSNR$\uparrow$ & SSIM$\uparrow$ & LPIPS$\downarrow$ & CLIP$\uparrow$ & Contextual$\downarrow$ \\
        \midrule
        \xmark & \xmark & 19.2491 & 0.8497 & 0.1361 & 0.7986 & 1.6399 \\
        \cmark & \xmark & 20.0198 & 0.8702 & 0.1187 & 0.8045 & 1.6378 \\
        \cmark & \cmark & \textbf{20.2452} & \textbf{0.8729} & \textbf{0.1117} & \textbf{0.8238} & \textbf{1.6031} \\
        \bottomrule
\end{tabular}
}
\caption{Quantitative ablation study on our test dataset. (I) denotes the feature interaction between the 2D Diffusion model and the reconstruction model, and (II) represents the injection of reference view features during the 2D Diffusion denoising process.}
\label{table:ablation}
% \vspace{-0.6cm}
\end{table}

\begin{figure*}
    \centering
    \includegraphics[width=0.8\textwidth]{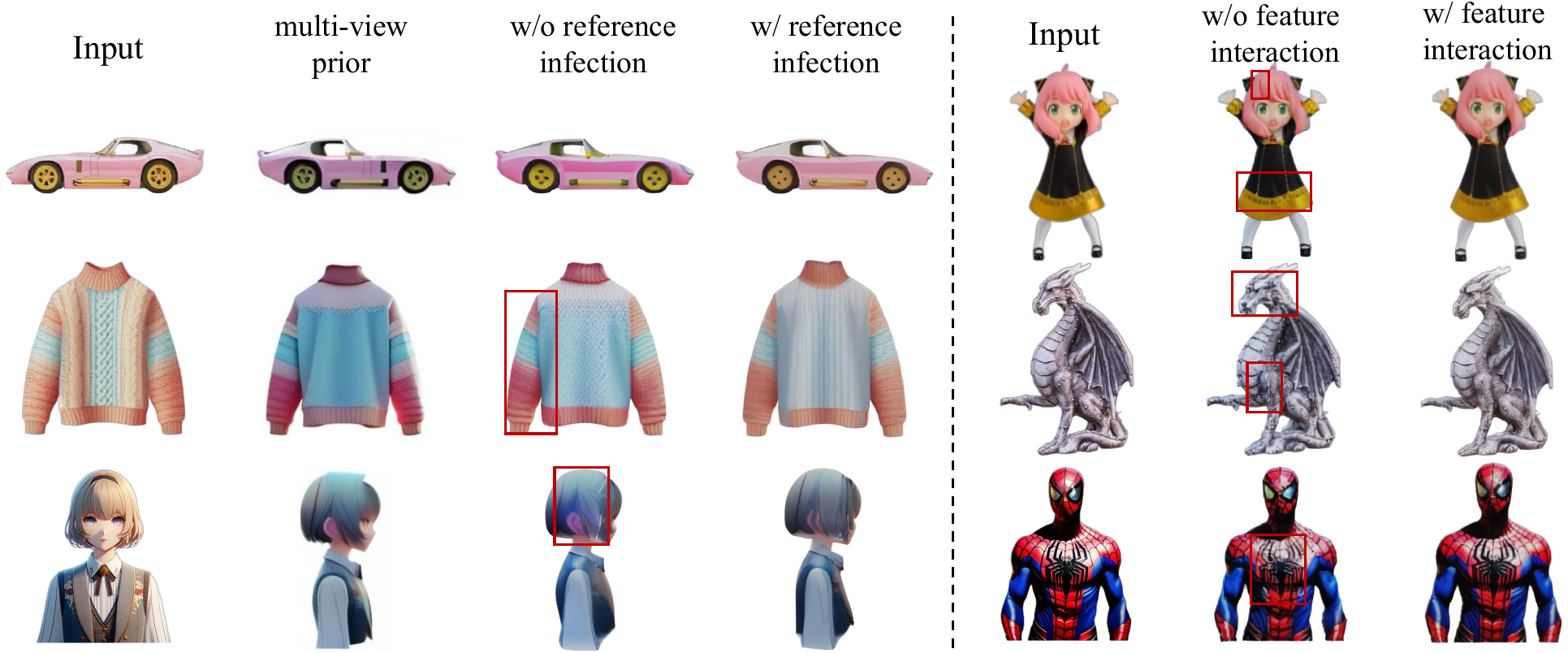}
    \caption{Qualitative ablation study by removing reference-view injection or feature interaction between 2D diffusion and reconstruction model. Multi-view prior refers to the multi-view images generated by the multi-view diffusion,  used as priors of 2D Diffusion model through DDIM inversion. The red boxes highlight some abnormal textures. Reference-view injection can reduce textures in the multi-view prior that are inconsistent with input, while the absence of feature interaction significantly degrades the reconstruction quality.}
    \label{fig:ablation}
\end{figure*}

\subsection{Ablation and Diverse Generation}
\begin{figure}
    \centering
    \includegraphics[width=0.45\textwidth]{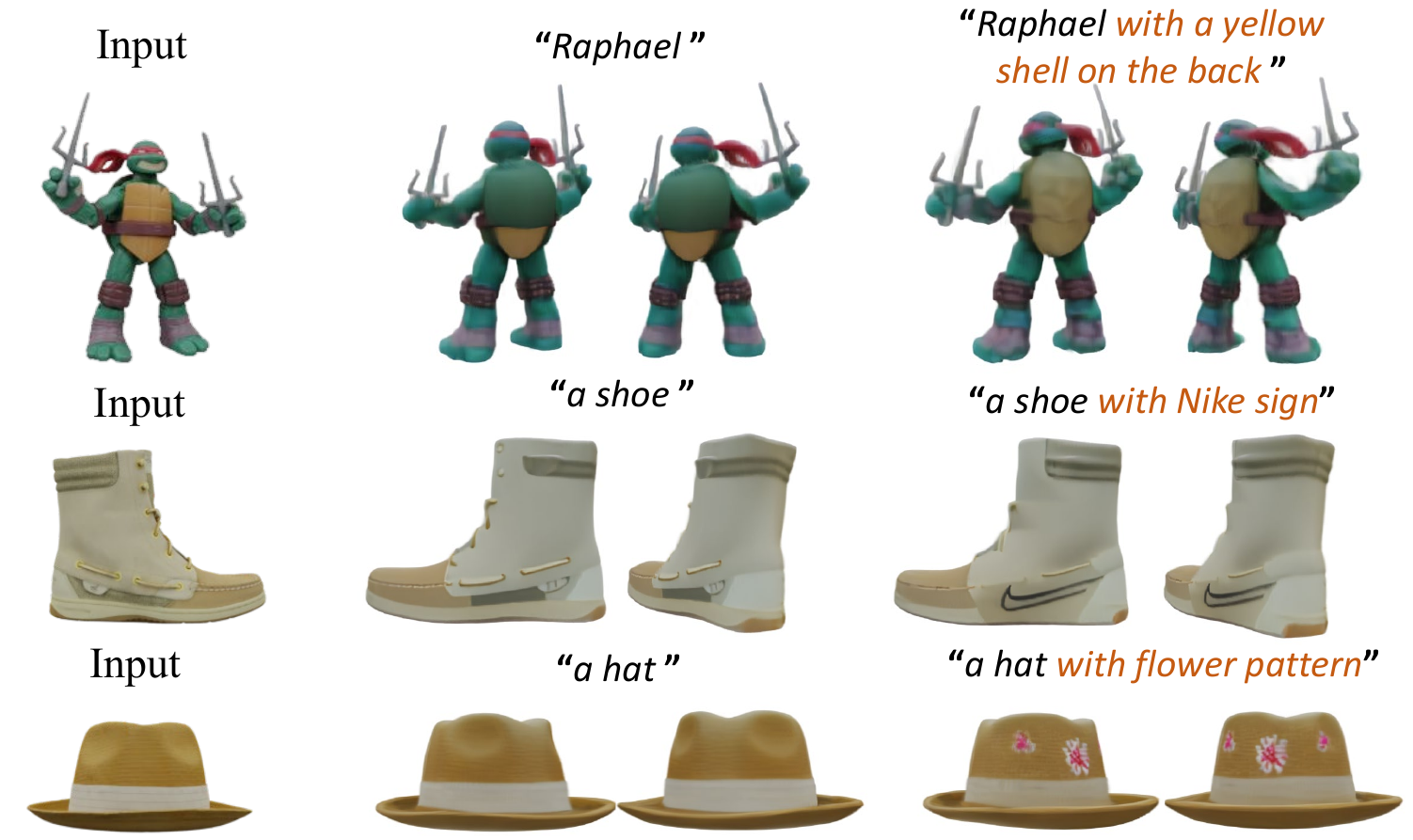}
    \caption{Diverse 3D generation. We can achieve controllable and diversified generation by using a variety of customized text.}
    \label{fig:text_control}
\end{figure}

In this section, we provide detailed quantitative and qualitative analysis, as shown in Figure~\ref{fig:ablation} and Table~\ref{table:ablation}. We also experimented with leveraging the text capabilities of the 2D diffusion model to control the generation of unseen regions from non-input viewpoints, as illustrated in Figure~\ref{fig:text_control}.

\textbf{Effectiveness of Feature Interaction.} The reconstruction model, trained only on a limited synthetic 3D dataset, often lacks the capability to accurately reconstruct complex and detailed textures in real-world scenarios, resulting in blurry textures, as depicted by the red boxes on the right side of Figure~\ref{fig:ablation}. The 2D diffusion model, trained on a large number of real web images, exhibits robust performance on real-world textures. The feature interaction between the encoder of the 2D diffusion model and the decoder of the reconstruction model significantly enhances the reconstruction of complex texture details, as evidenced by comparing the red-boxed areas in the last two columns of Figure~\ref{fig:ablation}.
Table~\ref{table:ablation} also demonstrates that feature interaction significantly enhances the quality of 3D reconstruction.

\textbf{Effectiveness of Reference-view Injection.} When multi-view diffusion generates multi-view images with unrealistic textures or textures that do not match the reference view, the 2D diffusion model, using the multi-view prior obtained through DDIM inversion, can still produce textures inconsistent with the reference view. As shown in the third column on the left side of Figure~\ref{fig:ablation}, although multi-view interaction through the reconstruction model alleviates texture inconsistency to some extent, the car's body, the coat, and the girl's hair and ear still exhibit abnormal textures due to the inconsistent multi-view prior generated by the multi-view diffusion, leading to discrepancies with the reference view. By injecting information from the reference view into the 2D diffusion denoising process, we can generate multi-view textures that are more consistent with the reference view, as shown in the fourth column in Figure~\ref{fig:ablation}. Table~\ref{table:ablation} also proves  reference-view injection can enhance the consistency of textures, as evidenced by increased CLIP similarity and reduced contextual distance.

\textbf{Diverse Generation.} Benefiting from the 2D diffusion model's excellent text alignment capability, we can apply diverse and customized text to control one or more non-input views, generating more varied textures in areas not visible from the reference view. As shown in Figure~\ref{fig:text_control}, the texture  in the second column are primarily based on the multi-view prior generated by the multi-view diffusion and the injection of reference view information during the denoising process. By incorporating fine-grained textual information as conditions, we can achieve diversified and customized 3D generation, as illustrated in the last column.

\begin{table*}
\centering
\resizebox{0.8\textwidth}{!}{
\begin{tabular}{lccccc}
\toprule
          Methods/Metrics &    PSNR$\uparrow$ &   SSIM$\uparrow$ &  LPIPS$\downarrow$ &   CLIP-Similarity$\uparrow$ &  Contextual-Dis$\downarrow$ \\
\midrule
% \textit{\textbf{Optimization-Based Methods}}\\
      
   DreamGaussian~\cite{tang2023dreamgaussian}& 18.2768 & 0.8342 & 0.1891 & 0.7488 &      1.2031 \\
        Wonder3D~\cite{long2024wonder3d} & 17.9891 & 0.8336 & 0.1877 & 0.7961 & 1.2607 \\
        \midrule
% \textit{\textbf{Feed-Forward Based Methods}}\\
One-2-3-45~\cite{liu2024one}& 16.0064 & 0.8186 & 0.2453 & 0.6623 &      1.4687 \\
TriplaneGaussian~\cite{zou2024triplane}& 17.9614 & 0.8404 & 0.1889 & 0.7765 & 1.2258\\
         OpenLRM~\cite{hong2023lrm} &    18.3686 & 0.8377 & 0.1733 & 0.8203 & 1.0861 \\
         LGM~\cite{tang2024lgm} & 19.4269 & 0.8539 & 0.1395 & 0.8389 & 0.9314\\
            Cycle3D (Ours) &\textbf{21.4841} & \textbf{0.8845} & \textbf{0.1155} & \textbf{0.8583} & \textbf{0.8346} \\
\bottomrule
\end{tabular}
}
\caption{We show quantitative results of image-to-3D in terms of  PSNR$\uparrow$ / SSIM$\uparrow$ / LPIPS$\downarrow$ / CLIP-Similarity$\uparrow$ / Contextual-Distance$\downarrow$ for the evaluated GSO~\cite{downs2022google} dataset. The \textbf{bold} reflects the best result for optimization-based methods and feed-forward methods.}
\label{table:gso}
% \vspace{-0.6cm}
\end{table*}
\section{Limitations}

Due to the lack of large-scale 3D scene datasets, our current method is limited to object-level 3D generation and cannot be extended to scene-level generation.  when large-scale scene datasets become available in the community, future work can explore more complex 3D scene generation. 
\section{Conclusion}
In this paper, we introduce Cycle3D,  an image-to-3D generation framework that cyclically utilizes 2D diffusion-based generation model and the 3D reconstruction model during the multi-step diffusion process. 
% cascades the 2D diffusion model with reconstruction model into a unified diffusion process. 
As the denoising evolves, the 2D diffusion model progressively generates multi-view images with higher quality, while the reconstruction model gradually corrects 3D inconsistencies. 
% ultimately achieving high-quality and consistent generation. 
2D diffusion model can also control the generation of unseen views and inject reference-view information during denoising.  Reconstruction model further interacts with 2D diffusion, enhancing the reconstruction capability. 
Extensive experiments demonstrate that our method surpasses existing state-of-the-art baselines in  generation quality and consistency.

{
    \small
    \bibliographystyle{ieeenat_fullname}
    \bibliography{main}
}

% \input{sec/2_relatedwork}
% \input{sec/3_pre}
% \input{sec/4_method}
% \input{sec/5_experiment}
% \input{sec/6_discussion}
% \input{sec/6_conclusion}

\clearpage
\setcounter{page}{1}
\maketitlesupplementary

\def\rvx{{\mathbf{x}}}

\section{Implementation Details}
\textbf{Training.} As described in Section~\ref{details}, we used 8 NVIDIA A100 (80G) GPUs to train our Cycle3D with a batch size of 8. Each batch contains four multi-view input images $\rvx_{0}$
  and eight additional images to supervise the fine-tuning process as the Eq~\ref{eq:loss1}. During training, we employed the DDPM Scheduler~\cite{ho2020denoising} with a maximum diffusion step 1000 to sample noisy multi-view images 
$\rvx_{t}$. The text prompt was set to empty with a 30\% chance during training, and the condition image of the reference view was set to either noisy (like other multi-view images) with a 30\% probability or clean at timestep 0 with a 70\% probability. The entire training processes are detailed in Algorithm~\ref{alg:training}.

\textbf{Inference.} For sampling process, we used the DDIM scheduler~\cite{song2020denoising} and set the number of sampling steps to 25 in our experiments. In the 2D diffusion denoising process, we also utilize reference-view injection to improve the texture consistency of the 3D generation. The entire sampling processes are detailed in Algorithm~\ref{alg:sampling}.

\textbf{Mesh Extraction. } We follow the method~\cite{tang2024lgm} to extract meshes from the generated 3D Gaussians.

\section{Evaluation Metrics}
We use PSNR, SSIM, LPIPS~\cite{zhang2018unreasonable}, as well as CLIP similarity~\cite{radford2021learning} and contextual distance~\cite{mechrez2018contextual} to evaluate the quality of image-to-3D generation. For our collected test dataset, due to the lack of multi-view ground truth, we use PSNR, SSIM, and LPIPS  to measure pixel-level and perceptual generation quality at the reference view. Additionally, we use CLIP similarity and contextual distance to assess the consistency between the novel views and the reference view. For the GSO dataset~\cite{downs2022google}, which has multi-view ground truth, we calculate PSNR, SSIM, LPIPS, CLIP similarity, and contextual distance for each rendered view and its corresponding ground truth to evaluate the multi-view consistency of 3D generation.

\section{Additional Visual Results}
\begin{figure*}[!t]
    \centering
    \includegraphics[width=0.91\textwidth]{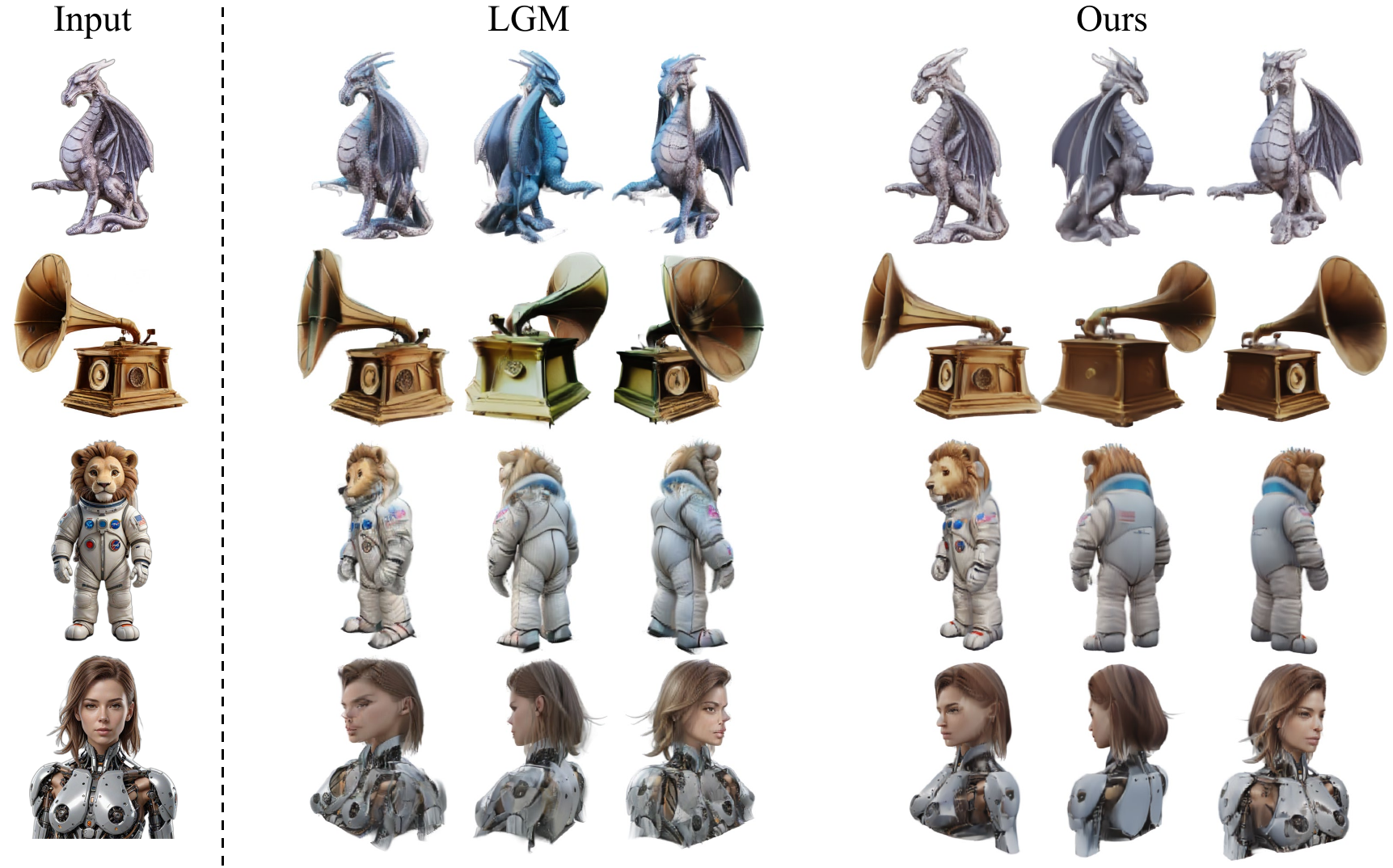}
    \caption{Visual comparison of more views between LGM~\cite{tang2024lgm} and our Cycle3D. Zoom in for more details. }
    \label{fig:supple_compare}
\end{figure*}

As shown in Figure~\ref{fig:supple_compare}, we compared the generation results of LGM and our Cycle3D across multiple viewpoints, demonstrating that our method achieves high-quality and consistent image-to-3D generation. Additionally, for convenience, we have provided more rendered videos on our website \url{https://pku-yuangroup.github.io/Cycle3D/}, which verify the superiority of our method compared to other existing baselines.

\begin{algorithm}[H]
  \caption{Training} \label{alg:training}
  \small
  \begin{algorithmic}[1]
  \Require Dataset of multi-view images $\rvx_{0}$ with the corresponding pose $\pi$,  a input image $\rvx^{\text{input}}$, text description $y$
  \Ensure Freeze pre-trained 2D diffusion model
  $\epsilon_{\theta}$ and optimize reconstruction model $\mathcal{G}_{\phi}$
    \Repeat
      \State $t \sim \mathrm{Uniform}(\{1, \dotsc, T\})$; $\boldsymbol{\epsilon} \sim \mathcal{N}(\mathbf{0},\mathbf{I})$
      \State $\mathbf{x}_{t} = \sqrt{\bar\alpha_t} \rvx_{0} + \sqrt{1-\bar{\alpha}_t}\boldsymbol{\epsilon}$
      \State $\hat{\mathbf{x}}_0 = \frac{1}{\sqrt{\bar{\alpha}_t}}( \mathbf{x}_t-\sqrt{1-\bar{\alpha}_t} \boldsymbol{\epsilon}_{\theta}(\mathbf{x}_t,  y, t))$ 
      \State $\hat{g} = \mathcal{G}_{\phi}\left(\hat{\mathbf{x}}_0 , t, \mathbf{F}_{\text{2D}} \right) $ 
      \textcolor{gray}{// enhance reconstruction quality with features of 2D diffusion ${\epsilon}_\theta$ as $ \mathbf{F}_{\text{2D}}$ }
      \State $
      \hat{\mathbf{x}}'_0 = \text{GS-renderer}\left(\hat{g}, \pi \right)$
      \State Compute loss $\mathcal{L}_{total}$ (~\cref{eq:loss1}) 
      \State Gradient step to update $ \mathcal{G}_{\phi}$

    \Until{converged}
  \end{algorithmic}
  \label{algm:train}
\end{algorithm}

\begin{algorithm}[H]
  \caption{Sampling} 
  \label{alg:sampling}
  \small
  \begin{algorithmic}[1]
  \Require A input image $\rvx^{\text{input}}$ and text prompt $y$; pre-trained 2D diffusion model $\epsilon_{\theta}$ and fine-tuned reconstruction model $\mathcal{G}_{\phi}$
  \Ensure 3D Gaussian output $g$ of the input image $\rvx^{\text{input}}$
  
    \vspace{.05in}
    \State $\rvx_T \sim \mathcal{N}(\mathbf{0}, \mathbf{I})$
    \For{$t=T, \dotsc, 1$}
      \State $\hat{\mathbf{x}}_{0} = \frac{1}{\sqrt{\bar{\alpha}_t}}( \mathbf{x}_t-\sqrt{1-\bar{\alpha}_t} \boldsymbol{\epsilon}_{\theta}(\mathbf{x}_t, y, t))$ 
      \State  $\hat{g} = \mathcal{G}_{\phi}\left(\hat{\mathbf{x}}_{0}, t, \mathbf{F}_{\text{2D}} \right) $
      \State $\hat{\mathbf{x}}'_0 = 
   \text{GS-renderer} \left(\hat{g}, \pi \right)$
      \State $\rvx_{t-1} = \frac{\sqrt{\alpha_{t}}\left(1-\bar{\alpha}_\text{t-1}\right)}{1-\bar{\alpha}_{t}} \rvx_{t} + \frac{\sqrt{\bar{\alpha}_\text{t-1}} \beta_{t}}{1-\bar{\alpha}_{t}} \hat{\mathbf{x}}'_0 $ \textcolor{gray}{// correct multi-view's inconsistency with 3D consistent renderings }
    \EndFor
    \vspace{.06in}
    \State \textbf{return} $g =  \mathcal{G}_{\phi}\left(\hat{\mathbf{x}}_{0}, \mathbf{F}_{\text{2D}}, t=0\right) $
     \vspace{.015in}
  \end{algorithmic}
\end{algorithm}

\begin{figure}[h]
    \centering
    \includegraphics[width=0.495\textwidth]{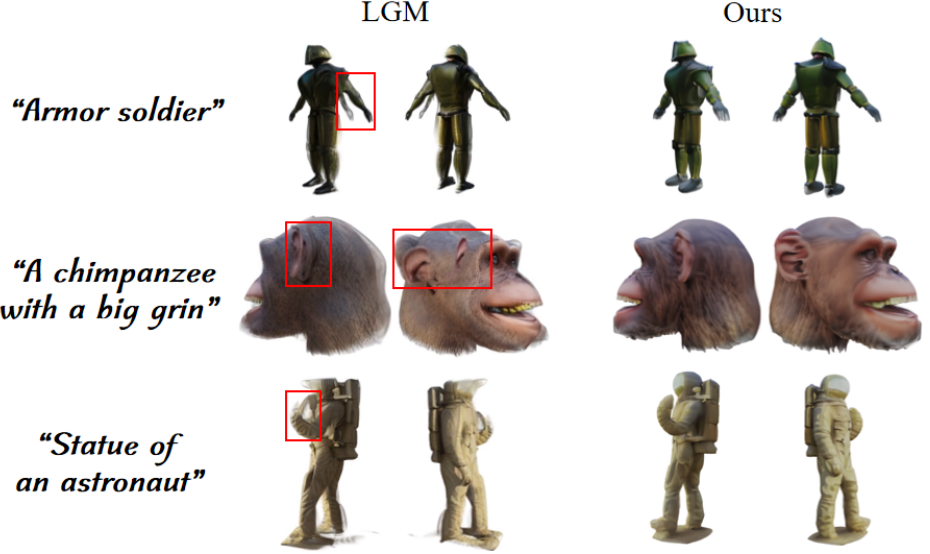}
    \caption{Qualitative comparisons on text-to-3D generation. The red boxes highlight some abnormal textures.  }
    \label{fig:supple_text}
\end{figure}

\section{Extension of Text-to-3D}
During training, the condition image of the reference view is occasionally noised along with other multi-view input images. This allows our Cycle3D to generate 3D objects conditioned by zero input view, i.e. text-to-3D. In the sampling process, the multi-view prior obtained through DDIM inversion~\cite{song2020denoising} serves as the initial noise and is denoised simultaneously using the 2D diffusion model. In contrast, during the image-to-3D sampling process, the reference view remains clean throughout the process. As shown in Figure~\ref{fig:supple_text}, the results generated by LGM exhibit geometric artifacts and blurry textures in the warrior's arm, the chimpanzee's ear, and the astronaut's hand due to the low quality and inconsistency of the multi-view images produced by the multi-view diffusion. In contrast, our Cycle3D method achieves satisfactory text-to-3D results through high-quality and consistent multi-cycle generation.

\end{document}